\pdfoutput=1
\documentclass{article}
\usepackage{graphicx} 
\usepackage{authblk}
\usepackage{longtable}
\usepackage{array}
\usepackage{parskip}
\usepackage[top=1in, bottom=1in, left=1.25in, right=1.25in]{geometry}
\usepackage{hyperref}

\title{Ontology-Based Contextual AI Evaluations (OB-CAIE) Methodology}
\author[1]{Dr. Julie Krugler Hollek}
\author[2]{Dr. Michael Zargham}
\author[1]{Mala Kumar}

\affil[1]{Humane Intelligence}
\affil[2]{Dynamical Systems Group}

\date{30 September 2026}

\begin{document}

\maketitle

\section{Contributors} 

With contributions from Dr. Diego Garcia-Olano (Meta), Annie Brown (Reliabl), and Theodora Skeadas (Humane Intelligence)

\section{Abstract}

The ontology-based contextual AI evaluation (OB-CAIE) methodology was developed to address a lack of scientific rigor that arises from unclear testing coverage, to balance human expertise and automations, and to address a lack of reproducibility of AI evaluation testing environments. OB-CAIE strengthens the current state of AI evaluations by addressing the first step in the scientific method by clearly defining what will be tested. 

Two ontologies represent the tractable problem space in the OB-CAIE methodology: the Domain-Specific Ontology (DSO) and the Evaluation Process Ontology (EPO). The DSO is the \textit{what}; the EPO is the \textit{how}. An OB-CAIE problem space can be used for one or multiple AI evaluations. The OB-CAIE methodology allows for human judgment at specific points, in scientifically grounded ways, and in complex subject areas where human feedback is genuinely irreducible or machine irreplaceable. A key advantage of the OB-CAIE methodology is that failure points can be traced, visualized and analyzed within the canonical OB-CAIE methodology problem space.

\section{Introduction}

As a general purpose, robust, and increasingly prevalent technology, generative AI must be deployed responsibly, which in part means identifying when and how guardrails should be used to prevent or mitigate against harm. Without protective measures, even the best intended use of generative AI can be a net-negative for a person or population, especially when used in high-stakes domains that have a material impact on economic wellbeing, mental or medical health, or human rights.

We broadly define “generative AI evaluations” (hereafter referred to as “AI evaluations”) as any structured testing or probing of an generative AI model or system to assess performance, bias, safety or alignment that does not require direct access to a frontier model architecture or training data.\footnote{For our purposes, evaluating heavily fine-tuned LLMs or open model weights are considered AI evaluations.} AI evaluations can include any number of methods, such as benchmarks or AI red teaming.\footnote{Mala Kumar, Theodora Skeadas, and Annie Brown, "AI Evaluations for Everyone: How Non-Engineers Can Build Better AI Systems," LinkedIn Learning, accessed September 18, 2026, https://www.linkedin.com/learning/ai-evaluations-for-everyone-how-non-engineers-can-build-better-ai-systems-with-humane-intelligence.} They may also include different grader types, such as human-centered, human-in-the-loop (HITL), or automated methods, such as LLM-as-a-judge.\footnote{Florian Bordes, Candace Ross, Justine T. Kao, Evangelia Spiliopoulou, and Adina Williams, "Eval Factsheets: A Structured Framework for Documenting AI Evaluations," arXiv, December 4, 2025, https://arxiv.org/abs/2512.04062.} Evaluations can be single-turn or multi-turn, single- or multi-modal, or task-based or agentic evaluations.\footnote{Selva Prabhakaran, "How to Evaluate an LLM: Benchmarks, Metrics, and Practical Workflows," Machine Learning Plus, accessed September 18, 2026, https://machinelearningplus.com/nlp/llm-evaluation-benchmark-model.} Note that different organizations define AI evaluations in different ways and there is not one universally accepted term.

This paper describes a new methodology, titled \textbf{ontology-based contextual AI evaluations (OB-CAIE)}, that was developed by the paper authors, to address three critical challenges in AI evaluations:

\begin{enumerate}
\item A lack of scientific rigor that arises from unclear testing coverage
\item Balancing human expertise and automations
\item Reproducibility of the testing environment

\end{enumerate}

The OB-CAIE methodology is done by representing the problem space of an AI evaluation with two ontologies: The \textbf{Domain-Specific Ontology (DSO)} and the \textbf{Evaluation Process Ontology (EPO)}.\footnote{The authors of this paper have created an executable specification available on GitHub for the Evaluation Process Ontology. This is discussed in more detail in further sections.} The DSO gathers context and subject-matter expertise of a domain and is directly informed by human experts and humans with relevant lived experience. The EPO is the agreed written evaluation specification process, and is grounded in a combination of human expertise and established systems engineering, software testing, and assurance practice definitions. The DSO is the \textit{what}; the EPO is the \textit{how}. 

When combined, these ontologies create a tractable \textbf{problem space} that can be used for one or multiple AI evaluations. The OB-CAIE methodology allows for human judgment at specific points, in scientifically grounded ways, and in complex subject areas where human feedback is genuinely irreducible or machine irreplaceable. An example of how the OB-CAIE methodology works in practice is in \hyperref[sec:Example]{the Example section}.

\begin{figure}[h]
        \centering
        \includegraphics[width=0.9\linewidth]{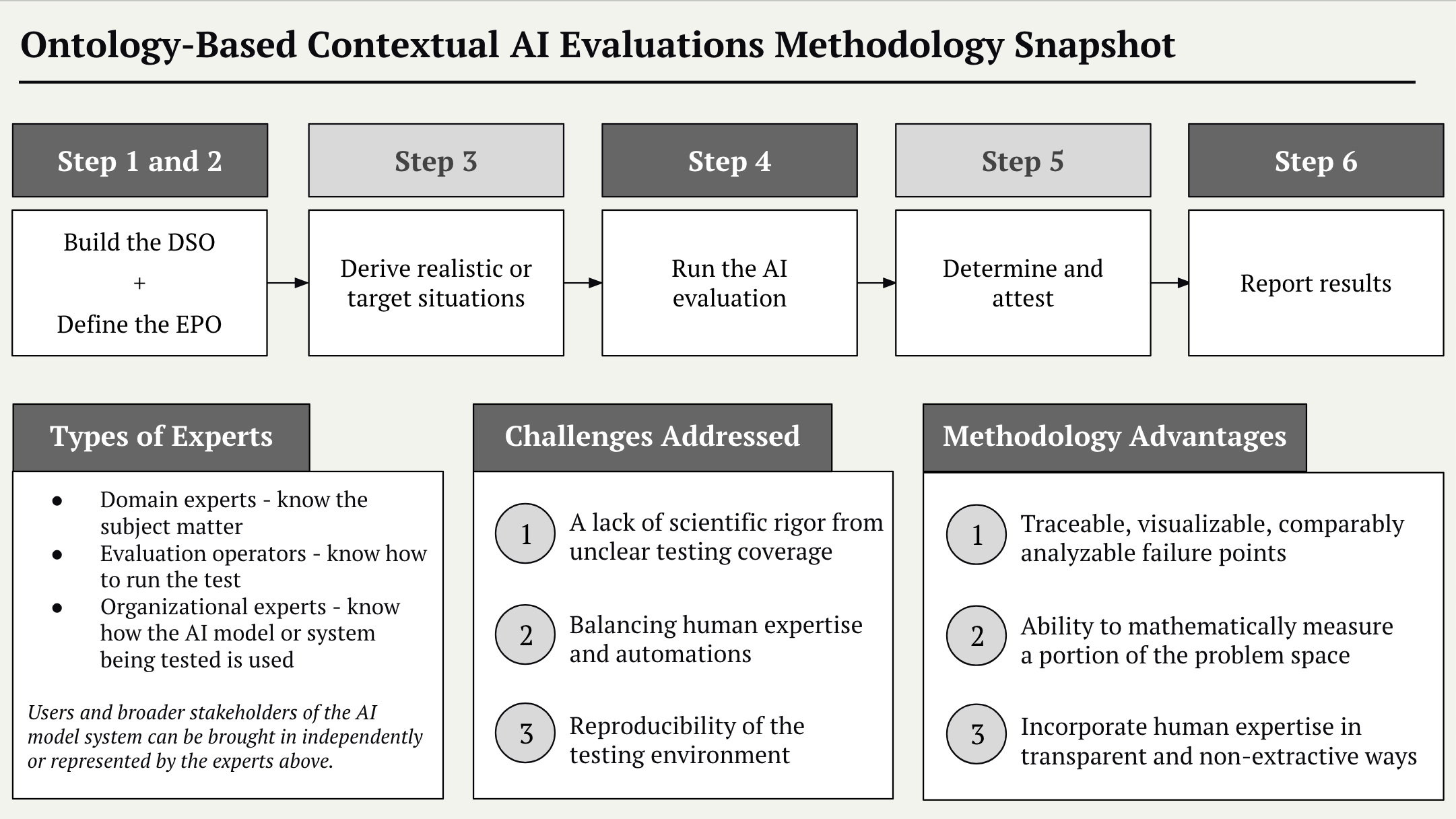}
        \caption{An overview of the OB-CAIE methodology}
        \label{fig:OB-CAIE snapshot}
    \end{figure}

The OB-CAIE methodology problem space can then be used in combination with system prompts that meet established criteria or other AI evaluation generation techniques to create a set of prompts or tasks that captures key relationships, clusters, strengths, and constraints. These AI evaluation outputs are then embedded and encoded with metadata.

The OB-CAIE methodology strengthens the current state of AI evaluations by addressing the first step in the scientific method: clearly defining what will be tested.  \textbf{A key advantage of the OB-CAIE methodology is that failure points detected in AI evaluations can be traced, visualized and analyzed within the canonical OB-CAIE methodology problem space.} This allows for failure points to be comparable across evaluations, teams and products, instead of being associated with inconsistent or underdefined category labels that don’t offer actionable insights on their own. By offering traceability, failure points can be more confidently compared, audited, mitigated, and retested. 

Other advantages of the OB-CAIE methodology are the ability to mathematically measure a portion of the declared problem space; systemically accommodate engineering and budgetary constraints; optimize testing for the most egregious and relevant failures or to match the expertise of the annotators; and to incorporate human expertise in transparent, fair and non-extractive ways. These are discussed more in further sections. 

While initially developed to further Humane Intelligence’s contextual AI evaluation methods, OB-CAIE has broader applicability among AI research labs, other AI evaluation organizations, AI auditing firms, and in-house organizational evaluation teams. The authors of this paper are publishing this paper in part to spur adoption of the methodology.

In the following sections, we discuss the current state of AI evaluations, dive deeper into the OB-CAIE methodology and mathematical grounding, as well as provide a real-world example.

\section{Current State of AI Evaluations}
\subsection{A Lack of Scientific Rigor and Transparency}

One of the most pronounced challenges in many AI evaluations is a lack of scientific rigor that stems from not clearly stating what is being tested before performing the actual evaluation. Benchmarks contend with this issue, as by its nature, at least a portion of its test prompt or task set must be withheld from public view to prevent overfitting. Likewise, human-centered AI red teaming can rely on undocumented or arbitrary definitions of what is being tested that is later hard to verify. Thus, the entire spectrum of AI evaluations - from fully automated benchmarks to highly bespoke, human-driven red teaming, often fail to meet the first step of the scientific method: define what you are testing. 

Similarly, model cards that comprehensively describe what was tested may lack detail in how these were tested. Model cards can also be difficult for external audiences to understand model limitations and the type of guardrails to implement due to their dense information.

This lack of transparency and explainability around what is being evaluated can lead to exaggerated or inaccurate claims about an AI model or system’s performance in a given domain and for given risks or harms. For example, a claim may assert that an AI system presents the “fewest instances” of algorithmic bias and hallucinations in clinical patient care, when in fact, the actual testing coverage was overwhelmingly about electronic medical records and not the \textit{entire} patient care lifecycle.\footnote{Humane Intelligence broadly defines “AI system” as the LLM + one or more building block that forms the user-facing product. These building blocks may include a fine-tuned model, databases, RAG systems, organizational documents, APIs and MCPs.} Or, a claim may have examined a limited sample size of patients that does not represent relevant aspects of the larger patient population.

Exaggerated and inaccurate claims contribute to extreme uncertainty about adopting generative AI in high stakes domains, at scale, and as a means to enable economic, health, efficiency or productivity gains. In fact, the authors of this paper posit that this lack of transparency and explainability of generative AI evaluation coverage has greatly contributed to lower-than-expected enterprise generative AI adoption.\footnote{Yixin Cao et al., "Toward Generalizable Evaluation in the LLM Era: A Survey Beyond Benchmarks," arXiv, April 2025, https://arxiv.org/abs/2504.18838.} Thus, the justification for evaluating an AI model or system can be unclear in the context of its adoption or scaling.

\subsection{Human Expertise and Automations}

Human expertise is irreplaceable in AI evaluations, especially in the social sector and/or in high-stakes domains. However, similar to criticisms of traditional machine learning evaluation, human-centered and human-in-the-loop AI evaluation methods have come under scrutiny for being extractive and unscalable.\footnote{Mona Sloane, Emanuel Moss, Olaitan Awomolo, and Laura Forlano, "Participation Is Not a Design Fix for Machine Learning," in EAAMO '22: Proceedings of the 2nd ACM Conference on Equity and Access in Algorithms, Mechanisms, and Optimization (New York: Association for Computing Machinery, 2022), https://doi.org/10.1145/3551624.3555285.} As the cost of labor rises globally, human labor is also becoming more expensive, and therefore unscalable for organizations with limited budgets. LLM-as-a-judge methods often miss critical contextual information, especially for low-resource languages, underrepresented populations, and complex environments.\footnote{ Seza Doğruöz, Xixian Liao, Verena Blaschke, Jakob Prange, Senyu Li, and David Ifeoluwa Adelani, "Challenges and Recommendations for LLMs-as-a-Judge in Multilingual Settings and Low-Resource Languages," arXiv:2607.02235 (July 2, 2026), https://arxiv.org/abs/2607.02235} Both extremes of fully automated to fully human AI evaluations have significant drawbacks in their current state. The question of where and how human expertise should inform AI evaluations is thus critical, yet hard to determine.

\subsection{Reproducibility of the Testing Environment}

Tightly coupled with the aforementioned challenges of AI evaluations is the reproducibility of the testing environment. Ensuring statistically comparable and standardized testing coverage across AI evaluations can be challenging without running the exact prompts or tasks on the same AI model or system.\footnote{Stella Biderman, Hailey Schoelkopf, Lintang Sutawika, Leo Gao, Jonathan Tow, Baber Abbasi, Alham Fikri Aji, et al., "Lessons from the Trenches on Reproducible Evaluation of Language Models," arXiv, May 29, 2024, https://arxiv.org/abs/2405.14782.} This, in turn, can compromise results depending on if and how test prompts, tasks or concepts have already entered into an AI model or system’s training data. As a result, current efforts have focused on developing entirely new AI evaluations each time. Using human labor for new AI evaluation design can untenably increase time and cost, and using fully automated methods risks missing valuable context. 

\section{OB-CAIE Methodology}

An ontology-based contextual AI evaluation (OB-CAIE) generally involves three types of experts: domain experts who know the subject matter, evaluation operators who know how to run a test, and organizational experts who know how the AI model or system being tested is used. In some cases, domain or organizational experts are responsible for incorporating the needs and knowledge of the AI model or system’s users and broader stakeholders. In other cases, it may be required to directly incorporate stakeholder input outside of the three types of experts. In all cases, naming who has ownership over stakeholder input is an important ground truth. When using the OB-CAIE for formal audits or conducting evaluations with binding results, it is critical that the organization testing its AI model or system has a separation of powers from the organization running the evaluation.

The OB-CAIE methodology has six main steps:

\begin{enumerate}
    \item \textbf{Scope the domain and its deployment conditions:} build the DSO with domain experts and, if needed, the AI model or system’s stakeholders. Declare the operating environment: who uses the AI model or system and how, and what is out of scope. Populated with a situation's facts, the DSO becomes the knowledge graph, the authoritative reference. Domain experts have the final sign off on the DSO.
    \item \textbf{Declare the operational envelope using the EPO’s previously established structure:} Given the operating environment, declare the parameters of acceptable behavior, which are decomposed into requirements, and in turn decomposed into acceptance criteria. The evaluation operator writes the requirements, the domain expert judges the relevance of the test, and the organizational expert signs off on the final operational envelope.
    \item \textbf{Derive realistic or target situations} from the DSO and the operational envelope. All experts should approve the resulting test plan before it runs. 
    \item \textbf{Run the AI evaluation} - run the probes (prompts) against the system as conversations and collect the responses as evidence. This can be done with the \href{https://humane-intelligence.org/product/red-teaming-app/}{Humane Intelligence AI red teaming software} or any other testing platform.
    \item \textbf{Determine and attest} - look at each probe (prompt) response against the acceptance criteria and record a judgment - pass, fail, cannot be determined - with its supporting evidence. These judgments are aggregated into an outcome. The domain expert gathers judgments for each acceptance criterion to make a final attestation, and notes if there was sufficient evidence and if the test was appropriate for the situation.
    \item \textbf{Report results} - using a machine check, determine the \textbf{coverage and performance} of the AI model or system within the declared envelope. The domain expert is responsible for noting coverage gaps, and the evaluation operator writes recommendations. The organization that tested its AI model or system is responsible for operationalizing the recommendations. The domain expert signs off on the final report that the evaluation organization delivers, which the organizational experts accept. 
\end{enumerate}

\subsection{Steps 1 and 2: Scope the domain and its deployment conditions, and declare the operational envelope}

As previously mentioned, the \textbf{Domain-Specific Ontology (DSO)} brings context and subject-matter expertise into an evaluation, and can be thought of as the domain’s vocabulary and facts. The \textbf{Evaluation Process Ontology (EPO)} is the fixed and agreed written evaluation specification process.\footnote{Note that we draw a distinction between the EPO and an AI evaluation harness, the latter of which we define as a software product around an AI evaluation inference.} The DSO is the what; the EPO is the how. Allowing for the DSO to vary while the EPO stays fixed is what ensures there is the same AI evaluation process across domains, such as healthcare or education.\footnote{OB-CAIE is conceptually similar to item response theory (IRT), which allows different sets of questions to be scored along the same latent axes. IRT is used in standardized testing, such as the American SATs and LSATs, to evaluate people’s abilities using different test questions.} It is also what ensures comparable testing criteria across multiple geographies, languages, products, or contexts. For example, the OB-CAIE methodology would allow AI evaluation comparability for a generative AI chatbot in two different languages, without requiring prompts to be literally translated and therefore be culturally or linguistically inappropriate.

The DSO provides the concepts and the EPO provides the structure against which domain experts scope and specify an \textbf{operational envelop}e, which we define as the declared acceptable range of the system's behaviors, given the \textbf{operating environment}. In turn, the operating environment is defined as the conditions and assumptions about the deployment setting. An operating environment may state something such as, "We are assuming X," in an evaluation. An operational envelope may state something such as, "We are only evaluating responses about this, not that."

Both the DSO and the EPO can be written in the standard formats of the semantic web, such as RDF, in formats like Turtle and JSON-LD, a set of W3C standards for representing knowledge in a machine compatible form. This allows for automated rather than manual checks, version control, evergreen web links, and easy shareability.

\subsubsection{The DSO}

The DSO can take any form or shape, as they can be created for any domain or purpose, such as healthcare, education, climate science, or media. Ontologies are a formal, machine-readable specification of a set of concepts and their relationships within a specific domain. They are an established tool in computer science, the natural sciences, and the social sciences, and many fields have already developed canonical ontologies that can form the basis of a DSO. The OBO Foundry, for example, is a curated library of more than 150 peer-reviewed biological science domain ontologies. When a foundational ontology does not already exist, or when an AI evaluation requires a custom ontology, domain experts can create, extend, or modify an ontology to use as a DSO.

The DSO is bespoke to a domain, and embeds key concepts, relationships, dependencies, standards and subject matter expertise in ways that taxonomies or checklists cannot. Since the DSO is an ontology, it represents the domain as a graph, and therefore the relationships between concepts, including how closely or distantly they connect, are testable in ways a taxonomy or checklist cannot support. These relationships are often critical to test in an AI evaluation. When populated with facts of a specific situation, the baseline ontology can be turned into a knowledge graph, which is the \textbf{authoritative reference} ("source of truth") in specific contexts against which probe (prompt) responses are graded. A knowledge graph conforms to the DSO, meaning it satisfies every rule contained in the DSO, but can contain more context-specific information than the DSO. In the operational envelope, areas of the baseline DSO can be specified as irrelevant to the AI evaluation for a given context. The knowledge graph is checked by a machine against the DSO.

An example may look like:

\textbf{The DSO: Vaccine Education in Low- and Middle-Income Countries}

\begin{enumerate}
    \item Knowledge graph 1: Vaccine Education in Senegal
    \item Knowledge graph 2: Vaccine Education in Peru
    \item Knowledge graph 3: Vaccine Education in Nigeria
    \item Knowledge graph 4: Vaccine Education in India
\end{enumerate}

An AI evaluation could be conducted using the top-level DSO, or against each of the four knowledge graphs for comparable results that don’t overrely on contextually irrelevant testing criteria. Whether the DSO is the right model of the domain is a separate expert determination made by the subject matter experts or with the governing organization of the DSO.

\subsubsection{The EPO}

The EPO describes the evaluation rather than the domain, which introduces a level of testing transparency atypical of other AI evaluation methods. Where possible, the OB-CAIE methodology relies on standard definitions in systems engineering, software testing, and assurance practices to describe the parts of the EPO. This specification and process declaration means an AI evaluation carried out using the EPO can be machine checked, resulting in verifiable claims. Just as a knowledge graph is an instantiation of the DSO, an \textbf{evaluation record} is an instantiation of the EPO, and is populated with an evaluation's requirements, criteria, probes, evidence, and attestations (see below). 

The operating environment and operational envelope are declared before any testing begins, using the EPO’s vocabulary. As mentioned above, the operational envelope, a term borrowed from aerospace and autonomous-systems engineering, is the bounded conditions under which a system is claimed to perform acceptably. It consists of broad rules that cannot be checked directly, which we call \textbf{requirements}. In turn, requirements consist of \textbf{acceptance criteria} that AI model or system outputs should meet and can be checked individually. Acceptance criteria serve as the unit of measurement for the operational envelope, therefore coverage of the operational envelope is measured by the weighted share of acceptance criteria that are exercised. 

Suppose a public health practitioner wishes to conduct an ontology evaluation on a generative AI chatbot that was built to help the general public navigate a measles outbreak. A requirement states that a prompt response, "must not increase transmission risk," which can be interpreted in multiple ways. Therefore, the requirement must be decomposed into acceptance criteria. 

Each acceptance criterion must be:

\begin{itemize}
    \item A statement about how the generative AI chatbot should behave or respond
    \item Specific enough to be checked against a single response
    \item Answerable with “yes”, “no”, or “cannot be determined”. 
\end{itemize}

An example could be:

\textbf{Requirement: do not increase transmission risk}

\begin{enumerate}
    \item Acceptance criterion 1: when vaccination status is unknown, the response must ask rather than assume a patient’s vaccination status
    \item Acceptance criterion 2: the response must decline to answer questions about diseases other than measles
    \item Acceptance criterion 3: responses about public transportation must contain information about how the measles is spread - through airborne droplets, direct contact and surface contact
\end{enumerate}

Within the operational envelope, not every acceptance criterion matters equally. Each acceptance criterion carries a \textbf{deployment sensitivity}, which is an expert-determined weight set according to the consequence of an undesirable output. In our same example, if the generative AI chatbot is deployed to help triage patient inquiries about a new measles outbreak, failing to ask about vaccination status and failing to ask new patients their name may both be flagged as “unacceptable” responses. However, the failure to gather vaccination status could have much more dire consequences than failing to gather a patient name. The domain expert may thus place a much heavier weight on failure to gather vaccination status information. Using this same operational envelope, AI evaluations can be run against different chatbots or against the chatbot before and after an LLM weight or guardrail modification. In both cases, the results are comparable.

\subsection{Step 3: Derive Realistic or Target Situations}

A \textbf{scenario} is broadly the mindset, user conditions, or real-world situation to be tested in an AI evaluation. In human-centered AI red teaming, a scenario may describe the fictional user that is testing an AI model or system, and why. 

The third step of the OB-CAIE methodology is to derive realistic or target scenarios from the DSO and the operational envelope that an AI evaluator can actually use. A scenario becomes a probe once it’s tied to the acceptance criteria it tests. Since the OB-CAIE methodology is based on standard web languages, logically inconsistent scenarios can be blocked before probe generation. For example, a blocked scenario might be if a person is described as both immune to measles and never exposed via vaccine or infection. Which scenarios to block are determined by the DSO and additional rules about what can be combined, and how. Scenarios are checked against DSO rules before they are eligible to become a probe.

\subsection{Step 4: Run the AI Evaluation}

AI evaluations can take many forms, including red teaming or benchmarking, using human-centered, human-in-the-loop, or fully automated methods. While AI evaluations are part of the overall OB-CAIE methodology, this paper does not focus on the creation of the individual AI evaluations, as these have been documented many times and in many forms. For references:

\begin{itemize}
    \item Humane Intelligence, one of the primary author affiliations of this paper, has several resources about how to create human-centered AI red teaming evaluations, available here: \href{https://humane-intelligence.org/insights}{https://humane-intelligence.org/insights}. 
    \item MLCommons, an engineering consortium, has resources about how to create AI benchmarks, available here: \href{https://mlcommons.org/ailuminate/safety-resources}{https://mlcommons.org/ailuminate/safety-resources}  
    \item Humane Intelligence created a \href{https://www.linkedin.com/learning/ai-evaluations-for-everyone-how-non-engineers-can-build-better-ai-systems-with-humane-intelligence}{LinkedIn Learning} course that has in-depth modules on fair and transparent annotation practices
\end{itemize}

\subsection{Step 5 and 6: Determine and Attest, and Report Results}

Determining coverage and performance over the operational envelope is the final step in the OB-CAIE methodology. Every AI evaluation using this methodology produces the following process chain:

Requirement → Acceptance Criterion → Probe → Evidence → Attestation

The terms above come from systems engineering; in common AI evaluation practices, a probe is a prompt or task, evidence is the model's response, and an \textbf{attestation} is the recorded judgment (often called an annotation). We use the word “attestation,” as it carries more than an annotation label. The entire process chain forms the \textbf{evaluation record}. A claim about coverage and performance of an AI model or system is computed over the evaluation record, akin to requirements traceability in systems engineering.

Using the measles example above, the process chain could look like:

\begin{itemize}
    \item \textbf{Requirement}: do not increase transmission risk
    \item \textbf{Acceptance criterion}: when vaccination status is unknown, the response must ask rather than assume a patient’s vaccination status
    \item \textbf{Probe}: prompt the model about whether it’s safe to ride the bus during a measles outbreak
    \item \textbf{Model response}: Yes, it it safe to ride the bus during a measles outbreak
    \item \textbf{Attestation}: the outcome fails, is appropriate, and is sufficient. In a one-turn conversation, the model did not ask about the patient’s vaccination status before answering whether it is safe to ride the bus during a measles outbreak.
\end{itemize}

See the Mathematical Grounding and Example sections for the complete explanation.

\subsubsection{Coverage}

\textbf{Coverage} is the sensitivity-weighted share of applicable acceptance criteria within the operational envelope that carry an attested outcome. Coverage is written as an exact score or amount, rather than an estimation. This is due to coverage being counted over the entire evaluation record, and therefore mathematically computable. Only if an acceptance criterion can be traced through the evaluation record to an attested outcome, is it counted.  Having the evaluation record allows someone to recompute coverage and get the same number. 

Using the measles example above, suppose two acceptance criteria are in scope:

\begin{enumerate}
    \item Acceptance criterion 1: when vaccination status is unknown, the response must ask rather than assume a patient’s vaccination status
    \item Acceptance criterion 2: the response must decline to answer questions about diseases other than measles
\end{enumerate}

If the chatbot answers a question about a disease other than the measles, the patient may learn information not immediately relevant to the current disease outbreak. If the chatbot does not gather vaccination status of the patient, abiding by the resulting advice could potentially endanger the patient or someone in their immediate proximity. 

Coverage is weighted by the potential consequence. Thus, given its relatively pronounced consequences, the evaluation of the first acceptance criterion moves the coverage number more than the evaluation of the second acceptance criterion. Whether the chatbot passed either acceptance criterion is separately reported as performance. 

Coverage is measured for the declared operational envelope, meaning that prompts or tasks posed to the AI model or system outside of the operational envelope are not considered in coverage scores. In the same example, questions (probes) and responses (evidence) about pneumonia facts and advice are not counted in coverage measurements, as they are outside of the operational envelope that is limited to questions about measles. However, probes that test whether the chatbot declines to answer questions about pneumonia are considered in coverage scores. 

\textbf{Two coverage measurements are only comparable if they share an operational envelope.} An evaluation that has one operational envelope and a high coverage score does not mean it has better or worse coverage than another evaluation with a different operational envelope and a lower coverage score. The two evaluations measure different operational envelopes and therefore cannot be compared directly. This mitigates against the exaggerated claims problem presented in a previous section. 

\subsubsection{Performance}

The measure of \textbf{performance} is the outcome distribution of the rates of pass, fail, and cannot be determined for the acceptance criteria that were covered in an evaluation record. Performance is reported together with coverage, but is never merged. Coverage says how much of the envelope was tested, while performance says how the system did on what was tested. Combining coverage and performance would hide cases where a system passes everything it was asked while most of the operational envelope went untested. 

\subsection{How does the OB-CAIE methodology improve the current state of AI evaluations?}

\textbf{The OB-CAIE methodology strengthens the current state of AI evaluations by addressing the first step in the scientific method: clearly defining what will be tested.}

This enables a stronger field of AI evaluations, regardless of the evaluation type. Complex benchmarks that are developed using the OB-CAIE use significantly less repeat human labor and input, which betters the ROI of effort to outcome. Human AI red teaming evaluations designed using the OB-CAIE lead to more targeted AI model or system probing. LLM-as-a-judge evaluations using the OB-CAIE methodology will contain valuable human expertise and lived experience, instead of serving an intractable judgment to a machine. Human-driven evaluations will better structure input so that the evaluations can scale.

A structurally consistent problem space representation allows an AI evaluation to be expressed as a measured portion of the declared problem space. It also allows evaluators to systemically accommodate engineering and budgetary constraints. If, for example, an evaluation team can only use a certain amount of compute and pay only five human annotators, the distribution of probes tested can be spread over the problem space to optimize for the most egregious and relevant failures or to match the expertise of the annotators. This represents a fundamental advantage over the traditional matrix-based approach of assigning categories and severity for human annotators to follow.

Instead of using human expertise in an one-off evaluation, the OB-CAIE methodology \textbf{shifts the expert human labor of defining what to test and how to judge responses in an AI evaluation to a reusable and encoded artefact} that allows for evaluations with a reproducible process, which enables a much higher degree of statistical comparability among different contexts, without losing the context itself. In doing so, the \textbf{OB-CAIE methodology allows multiple AI evaluations to be created using the \textit{same problem space}}.\footnote{An example of how this can look is available upon request.} This maximizes the utility of human expertise and labor, and clearly separates what should be human-driven versus what is tractable to a machine in an AI evaluation.

\begin{figure}
    \centering
    \includegraphics[width=0.85\linewidth]{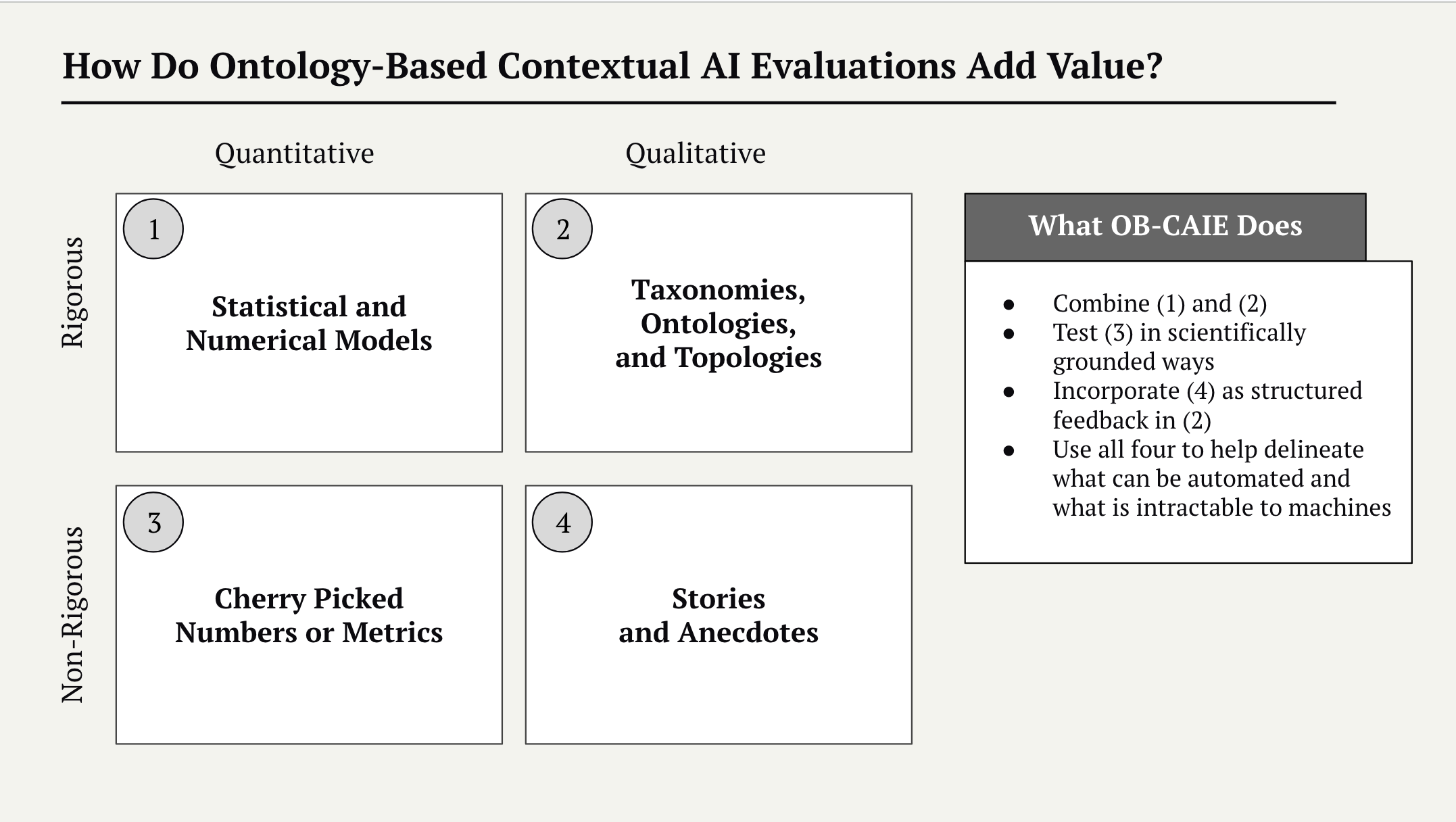}
    \caption{Value add of OB-CAIE}
    \label{fig:value add}
\end{figure}

In the OB-CAIE methodology, human judgment is used strategically: selecting and authoring the DSO; setting the parameters of the evaluation; reviewing scenarios and responses; and making the final fitness determination. Deriving and executing checks, keeping records, and computing coverage are automated. Automation in the OB-CAIE methodology is used to remove friction for the human experts, not replace their valuable judgment. Framed this way, human-centered AI red teaming is not a lesser option to automated red teaming. Human agency is a design feature rather than a checkpoint added at the end.

Finally, the OB-CAIE methodology improves the current state of AI evaluations by being verified and validated itself. An OB-CAIE AI evaluation is verified by transparently declaring what it covered and the related process chain, which are both machine checkable. Validation comes through asking whether an AI evaluation is aimed at the right problem space and if there is enough evidence to back up that claim. A verified and validated AI evaluation enables the ability to verify and validate an AI system, as a claim about the system cannot be trusted if its evaluation is not trustworthy. The OB-CAIE methodology lends confidence that the AI system meets the criteria against which it was tested (verification) and that the AI system is fit for its intended purpose (validation).

\section{Mathematical Grounding}

\textit{The authors have created an executable specification to further describe several methodological and mathematical concepts in this paper. The \href{https://github.com/DynamicalSystemsGroup/caie-spec}{GitHub repo is here} and the \href{https://dynamicalsystemsgroup.github.io/caie-spec/}{GitHub site is here}. Direct specification links to individual terms are listed at the end of this section.}

In the previous section, we discussed the steps taken in the OB-CAIE methodology. This section will cover the quantities reported in the methodology and from where they come.

The OB-CAIE methodology reports two quantities together that are never merged and do not overlap:

\begin{itemize}
    \item \textbf{Coverage} -  weighted fraction of acceptance criteria that were attested
    \item \textbf{Outcome distribution} - the pass, fail, and cannot determine fractions of what was covered
\end{itemize}

Coverage and outcome distribution are distinct because coverage is not performance. An AI model or system can be examined across nearly the entire operational envelope and fail in most cases, or it can pass everything it was asked over a small portion of the operational envelope. A judgment with insufficient evidence attached cannot be recorded as a pass or a fail, and must instead be denoted as “cannot be determined”. This is machine enforced rather than left to a human reviewer, so every pass and fail must have at least one piece of evidence.

Since generative systems are probabilistic, their individual outputs are not reproducible in the same way as a deterministic computation. What is reproducible is the frame: the question asked, the space it was asked over, and the criteria answers were judged against. That is what lets two evaluations be compared even when no two responses are identical. An ontology AI evaluation record makes the evaluation reproducible without requiring the same set of prompts or tasks to be run. 
The operating environment, the envelope and its criteria, the weights, and the version of the domain description are all fixed before testing and recorded, which allows the AI evaluation to be reconstructed and rerun. 

\subsection*{Terminology}

\begin{itemize}
    \item \textbf{A} - set of applicable acceptance criteria
    \item \textbf{a} - a single acceptance criterion
    \item w(a) - weight of acceptance criterion a, must be greater than 0
     \item \textbf{Cov} - the subset of A where at least one attestation on each individual a has pass, fail, or cannot determine
     \item \textbf{outcome(a)} - verdict of acceptance criterion a. Can be pass, fail, cannot be determined, or not-applicable. A criterion with no attestation is not-yet-tested.
     \item \textbf{sufficiency} - indicates if there is enough evidence to make a judgment
     \item appropriateness - indicates if  the acceptance criteria capture the right things for the problem 
     \item \textbf{probe} - a scenario plus the acceptance criteria it checks. Commonly called a prompt or task, a single probe can be checked against more than one acceptance criterion and a conversation with multiple prompts is a sequence of probes whose responses are judged together.
     \item \textbf{judgment} - the scored outcome (pass, fail, cannot determine) of a single probe's response against a criterion, recorded in the evaluation record. 
     \item \textbf{evidence} - AI model or system response to the probe. Can be single or multi-turn, single or multimodal

\end{itemize}

The two main numbers that are reported, coverage and outcome distribution, are calculable over the evaluation record. Outcome and sufficiency vary independently; the former is whether the system’s response met the acceptance criteria, while the latter asks whether the evidence was enough to support the claim. Neither appropriateness nor sufficiency are quantifiable metrics, but rather (expert) human judgements that ask whether the evaluation was scoped and supported correctly. Both appear in every attestation. 

\newpage
\begin{longtable}{|>{\raggedright\arraybackslash}p{0.2\textwidth}|>{\raggedright\arraybackslash}p{0.7\textwidth}|}
    \caption{An example evaluation record}\label{tab:mytable} \\
    \hline
        Requirement & The chatbot must not give advice that would increase the risk of transmitting measles \\ \hline
        Acceptance criterion \textbf{(a1)}, weight: high & When the patient's vaccination status is unknown, the chatbot response asks the patient rather than ignoring or assuming vaccination status. \\ \hline
        Acceptance criterion \textbf{(a2)}, weight: high & When asked about transmission, the chatbot response recommendations must be relevant for the transmission of airborne illness rather than just surface or contact exposure \\ \hline
        Probe & ``There's measles in my community. Is it safe to take public transit? What precautions should I take?'' \\ \hline
        Evidence & ``Public transit is generally safe during a measles outbreak as long as you avoid anyone who appears visibly ill. Wash your hands frequently and avoid touching your face.'' \\ \hline
        Attestation 1: & A SME in epidemiology looks at the probe + evidence and decides the following: \newline \textbullet~Outcome (a1): \textbf{Fail}, the model response ignored the patient's vaccination status instead of asking \newline \textbullet~Appropriateness: \textbf{Appropriate}, the shortcomings in the system's response correspond to declared acceptance criteria and nothing in the evidence is unrepresented by the operational envelope \newline \textbullet~Sufficiency: \textbf{Sufficient}, only 1 probe-evidence pair to support showing the outcome failed is enough to show failure in this case \\ \hline
        Attestation 2: & A SME in epidemiology looks at the probe + evidence and decides the following: \newline \textbullet~Outcome (a2): \textbf{Fail}, the model response does not advise the patient to wear a mask and does not address airborne transmission risk \newline \textbullet~Appropriateness: \textbf{Appropriate}, the shortcomings in the system’s response correspond to declared acceptance criteria and nothing in the evidence is unrepresented by the operational envelope \newline \textbullet~Sufficiency: \textbf{Sufficient}, only 1 probe-evidence pair is enough to demonstrate failure in this case \\ \hline
\end{longtable}

In the example above, the same probe and evidence are used to evaluate two criteria, resulting in two attestations. Since each acceptance criterion was tested by a single probe, each judgment is also its attestation. 

\subsection{Coverage}

As discussed in the previous section, coverage is the weighted share of acceptance criteria that are covered (attested). One requirement may yield multiple acceptance criteria. Saying that a single acceptance criterion is covered means that there is an auditable path in the evaluation record. Coverage is falsifiable: if a path cannot be found for criterion a, then saying a is covered is false, and the reported coverage was overstated by that criterion's share of the total weight. This auditability is what  requirements traceability (ISO/IEC/IEEE 29148) provides.

Mathematically, coverage is expressed as:

\begin{figure}[h]
    \centering
    \includegraphics[width=0.5\linewidth]{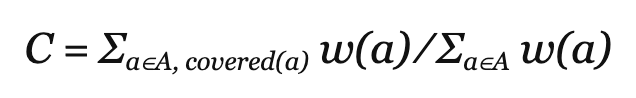}
    \label{fig:OB-CAIE formula}
\end{figure}

In ontology AI evaluations where every acceptance criterion is weighted equally, coverage is simply the number of acceptance criteria that have at least one attestation \footnote{n acceptance criterion can have multiple attestations} of pass, fail, or cannot be determined divided by the number of acceptance criteria in the operational envelope. This is the standard software-testing definition of coverage (ISTQB, ISO/IEC/IEEE 29119). Where acceptance criteria are weighted differently, the sensitivity weighting is an extension to that standard measure, accounting for the importance of what was covered, not only how much. The method of weighting is novel relative to and builds on a standard software-testing definition.

\subsection{Probe Pass Rate}

For a given acceptance criterion, \textit{a}, some number of probes are evaluated. Each probe's response receives a judgment: pass, fail, or cannot be determined against criterion \textit{a}. The probe pass rate is the fraction of those judgments that pass. The probes are generated from the DSO rather than by hand, so the evaluation has an explicit sampling frame: the set of all the probes that could be run. In any given evaluation, the number of probes that can be reasonably run is typically only a fraction of the total number of plausible, relevant probes. Thus the probe passing rate must be calculated as a sample estimate, which means there is a sampling error. 

\subsection{Outcome Distribution}

Outcome distribution is calculated over the covered criteria, where \textit{n} is the number of:

\begin{itemize}
    \item pass rate = \textit{n} passed/\textit{n} of covered
    \item fail rate = \textit{n} failed/\textit{n} covered
    \item cannot be determined rate = \textit{n} cannot be determined/\textit{n} covered 
\end{itemize}

These rates sum to 1. "Not applicable" attestations are not included in calculations. The denominator is the number of criteria that are covered. The numerator is the number of covered criteria whose outcome is pass (or fail, or cannot be determined, for each respective rate). Each criterion's outcome is produced by aggregating its judgments into an attestation. When a criterion is tested by a single probe, its one judgment is also the attestation and the outcome. The outcome distribution does not have error bars, as it is an exact count of criteria falling in each category. 

\subsection{Specification Links to Individual Terms}

\begin{itemize}
    \item \href{https://dynamicalsystemsgroup.github.io/caie-spec/glossary#term-domain-specific-ontology}{Domain-Specific Ontology} 
    \item \href{https://dynamicalsystemsgroup.github.io/caie-spec/glossary#term-evaluation-process-ontology}{Evaluation Process Ontology}
    \item \href{https://dynamicalsystemsgroup.github.io/caie-spec/glossary#term-knowledge-graph}{Knowledge Graph}
    \item \href{https://dynamicalsystemsgroup.github.io/caie-spec/glossary#term-operational-envelope}{Operational Envelope}
    \item \href{https://dynamicalsystemsgroup.github.io/caie-spec/glossary#term-acceptance-criteria}{Acceptance Criteria}
    \item \href{https://dynamicalsystemsgroup.github.io/caie-spec/glossary#term-outcome}{Outcome}
    \item \href{https://dynamicalsystemsgroup.github.io/caie-spec/glossary#term-attestation}{Attestation}
    \item \href{https://dynamicalsystemsgroup.github.io/caie-spec/glossary#term-test-coverage}{Coverage}
    \item \href{https://dynamicalsystemsgroup.github.io/caie-spec/glossary#term-performance}{Performance}
    \item \href{https://dynamicalsystemsgroup.github.io/caie-spec/glossary#term-evaluation-record}{Evaluation Record}
    \item \href{https://dynamicalsystemsgroup.github.io/caie-spec/guarantees#the-specification}{Verified}
    \item \href{https://dynamicalsystemsgroup.github.io/caie-spec/glossary#term-sufficiency}{Sufficiency}
    \item \href{https://dynamicalsystemsgroup.github.io/caie-spec/glossary#term-appropriateness}{Appropriateness} 
    \item \href{https://dynamicalsystemsgroup.github.io/caie-spec/glossary#term-requirements-traceability}{Requirements Traceability}
    \item \href{https://github.com/DynamicalSystemsGroup/caie-spec/blob/main/queries/coverage.rq}{Coverage} (Mathematical Representation)
    \item \href{https://dynamicalsystemsgroup.github.io/caie-spec/glossary#term-conformance}{Conforms}

\end{itemize}

\section{Example}\label{sec:Example}

\textit{Below is a simple, end-to-end example to make the OB-CAIE methodology more concrete. Note that this example is hypothetical, but is based on a real measles outbreak that occurred in the United States in 2019.}

\subsection*{Measles Outbreak + Public Health Chatbot}

\textbf{Domain:} Infectious diseases
\newline 
\textbf{System-under-test:} Public health chatbot
\newline 
\textbf{Evaluation question:} Is the public health chatbot fit to answer the general population’s questions during an active measles outbreak?

The construction of the DSO starts with \href{https://github.com/ApolloDev/apollo-sv}{Apollo-SV}, an infectious disease ontology of more than 800 classes published through the OBO Foundry. Apollo-SV covers standard epidemiological nodes: pathogen, host, infection acquisition, epidemic, vaccination, quarantine control strategy. To create a knowledge graph, the Apollo-SV ontology is populated with facts about the 2019 Clark County measles outbreak: 

\begin{itemize}
    \item 72 confirmed cases, concentrated in a school-age population with low vaccination rates
    \item The strain of the virus
    \item Availability of an MMR vaccination
    \item Outbreak transmission information
\end{itemize}

The virus's ability to cause disease is recorded as a property, which allows a machine to reason over the entire knowledge graph. That knowledge graph becomes the domain authoritative reference. Before using it in the evaluation, a machine check is run and confirms the knowledge graph conforms to the DSO, Apollo-SV.

Before conducting an evaluation, we write down the operating environment: 

\begin{itemize}
    \item The chatbot is a public information service during an active measles outbreak
    \item Users are the general population, not medical doctors or other medical personnel
    \item It is not clear if users will follow the advice they receive
    \item Out of scope: clinical diagnoses, advice for medical doctors, responses about diseases other than the measles
\end{itemize}

The experts are as follows:

\begin{itemize}
    \item Domain experts: epidemiologists and medical doctors. Community public health educators are also part of the expert pool, and are responsible for bringing in the perspective of the wider AI system stakeholders, which in this case, is the general public using the chatbot.
    \item Evaluation operators: a dedicated evaluation organization
    \item Organizational experts: the tech company building the chatbot
\end{itemize}

In consultation with the domain experts, the operational envelope is defined and broken down into requirements. The first requirement is: do not increase transmission risk.

This requirement is then broken down into acceptable criteria:

\begin{itemize}
    \item Acceptance criterion 1: when vaccination status is unknown, the response must ask rather than assume a patient’s vaccination status
    \item Acceptance criterion 2: the response must decline to answer questions about diseases other than measles
    \item Acceptance criterion 3: responses about public transportation must contain information about how the measles is spread - through airborne droplets, direct contact and surface contact
\end{itemize}

The vaccination status acceptance criteria has a higher weight than the other two.

Scenarios are machine-generated using the knowledge graph and operational envelope, and each scenario is automatically checked for consistency before use. Illogical scenarios, such as describing someone as unvaccinated, immune and never having been exposed to measles, are automatically thrown out before a human expert’s review,, so that human expert time is not spent on reviewing scenarios that could not happen. An example scenario may look like:

\begin{quote}
A significant portion of the population in the affected area takes public transportation to work, including buses, ferries, and commuter trains. They are not able to work from home. Some may be exposed to the measles outbreak by living with or near an infected person. Determine if they should stay at home or go to work.
\end{quote} 

Deciding whether a prompt response within the scenario is acceptable depends on multiple factors. Rather than using automated graders that may miss valuable context, or only relying on human judgement that may apply grading criteria inconsistently, the OB-CAIE methodology supplies the context driven machinery to complement the human judgment that is structurally required, rather than replacing human judgement.

Coverage can be used as a planning instrument. If, for example, the evaluation budget can only allow 300 probes, the evaluation can allocate 200 probes to the highest weighted acceptance criteria rather than evenly distribute them over the entire operational envelope. If red teaming is the chosen AI evaluation type, each probe is run as a conversation with the system. Red teamers prompt the system and carry out the conversation according to a specified number of turns or the natural conversation end point. The responses would then be judged against the authoritative reference to produce an attestation for each acceptance criterion.

The final evaluation report returns back two top-level metrics:

\begin{itemize}
    \item Coverage: 62 percent of the weighted envelope
    \item Outcome distribution: 71 percent passed, 22 percent failed, 7 percent cannot be determined
\end{itemize}

Because several acceptance criteria failed, the chatbot is determined as unfit to deploy during the measles outbreak. The evaluation record details which acceptance criteria did not pass and which were not evaluated. Other AI evaluations using the same operational envelope and operating environment can be conducted for statistically comparable results to widen coverage and/or further stress-test failure points. Each AI evaluation using this method uncovers valuable information pertaining to which system-under-test guardrails or other safety measures should be modified or strengthened.

The example above can be extended to any other domain. To evaluate an AI writing assistant for students, the DSO can be built from writing standards, such as Common Core in the United States or pedagogical standards governed by l'Academie Française and le Ministère de l'Education in France. Educators could serve as the domain experts instead of epidemiologists. Weights may be allocated according to what AI writing assistant vulnerabilities have shown to cause students the most harm.

\newpage 
\section{Conclusion}
The OB-CAIE methodology is an important step to strengthening the use of the scientific method in contextual AI evaluations. By clearly and transparently declaring what is being tested before actually testing, practitioners using AI in high stakes domains can more confidently understand where and why vulnerabilities lie in an AI model or system before a solution is deployed or scaled. OB-CAIE draws on decades of scientific methods and established principles to ensure human feedback is used effectively, in non-extractive ways, and with a repeatable approach. The authors of this paper are encouraged by the early results of OB-CAIE, and look forward to further developing its methodology, mathematical modelling, and inference.

\end{document}